\documentclass[a4paper, conference]{IEEEtran}

\usepackage{cite}
\usepackage{amsmath,amssymb,amsfonts}
\usepackage{algorithmic}
\usepackage{graphicx}
\usepackage{textcomp}
\usepackage{paralist}
\usepackage{eurosym}
\usepackage{flushend}
\usepackage{url}
\usepackage[utf8]{inputenc}
\usepackage[colorlinks=true]{hyperref}
\usepackage{soul}
\usepackage{caption}
\usepackage[list=true]{subcaption}
\usepackage{booktabs}
\usepackage{tabu}
\usepackage{wrapfig}
\usepackage{arydshln}
\usepackage[normalem]{ulem}
\usepackage{multirow}
\usepackage{siunitx}
\usepackage{adjustbox}
\usepackage{array}
\usepackage{diagbox}
\usepackage{multirow}
\usepackage{xcolor}

\begin{document}

\title{5G-DIL: Domain Incremental Learning with Similarity-Aware Sampling for Dynamic 5G Indoor Localization}

\author{\IEEEauthorblockN{Nisha L. Raichur\IEEEauthorrefmark{1},
    Lucas Heublein\IEEEauthorrefmark{1},
    Christopher Mutschler\IEEEauthorrefmark{1},
    \underline{Felix Ott}\IEEEauthorrefmark{1}}
    \IEEEauthorblockA{\IEEEauthorrefmark{1}Fraunhofer Institute for Integrated Circuits IIS, 90411 Nürnberg, Germany}
    \IEEEauthorblockA{\{nisha.lakshmana.raichur, lucas.heublein, christopher.mutschler, felix.ott\}@iis.fraunhofer.de}
}

%%%%%%%%%%%%%%%%%%%%%%%%%%%%%%%%% IEEE flag at the bottom %%%%%%%%%%%%%%%%%%%%%%%%%%%%%%%%%%%%%%%%%
\IEEEoverridecommandlockouts
\IEEEpubid{\makebox[\columnwidth]{
979-8-3315-1113-5/25/\$31.00~\copyright2025
IEEE \hfill} \hspace{\columnsep}\makebox[\columnwidth]{ }}

\maketitle

\begin{abstract}
Indoor positioning based on 5G data has achieved high accuracy through the adoption of recent machine learning (ML) techniques. However, the performance of learning-based methods degrades significantly when environmental conditions change, thereby hindering their applicability to new scenarios. Acquiring new training data for each environmental change and fine-tuning ML models is both time-consuming and resource-intensive. This paper introduces a domain incremental learning (DIL) approach for dynamic 5G indoor localization, called 5G-DIL, enabling rapid adaptation to environmental changes. We present a novel similarity-aware sampling technique based on the Chebyshev distance, designed to efficiently select specific exemplars from the previous environment while training only on the modified regions of the new environment. This avoids the need to train on the entire region, significantly reducing the time and resources required for adaptation without compromising localization accuracy. This approach requires as few as 50 exemplars from adaptation domains, significantly reducing training time while maintaining high positioning accuracy in previous environments. Comparative evaluations against state-of-the-art DIL techniques on a challenging real-world indoor dataset demonstrate the effectiveness of the proposed sample selection method. Our approach is adaptable to real-world non-line-of-sight propagation scenarios and achieves an MAE positioning error of 0.261 meters, even under dynamic environmental conditions. Code: \href{https://gitlab.cc-asp.fraunhofer.de/5g-pos/5g-dil}{https://gitlab.cc-asp.fraunhofer.de/5g-pos/5g-dil}
\end{abstract}
\IEEEpeerreviewmaketitle

\section{Introduction}
\label{label_introduction}

With the rapid development of location-based 5G services~\cite{floch_kacimi}, including smart homes, hospitals, warehouses, and shopping centers, indoor radio localization has emerged as a critical enabler~\cite{stahlke2022transfer,whiton_chen_tufvesson,alawieh_kontes,foliadis_garcia_gallacher,etiabi_njima,pan_wei_he,kerdjidj_himeur_sohail}. In such networks, the deployment of (multiple) base stations (BSs)~\cite{dai_chen_zhou} facilitates the collection of channel impulse response (CIR) over distributed links, which can be leveraged for precise positioning~\cite{foliadis_garcia}. CIR characterizes the channel across both spatial and frequency domains, benefitting from the substantial number of antennas and wide bandwidth available in current and future communication networks, such as 5G and 6G~\cite{chen_sarieddeen,vuckovic_hosseini}. Conventional radio-based positioning techniques are typically model-based and adopt a two-step approach. Initially, with CIR estimated at one or more BSs, parameters such as path delay, angle of arrival, reference signal receive power, and time difference of arrival are extracted. These measurements are then used to compute the position. However, these range-based methods often suffer from limitations due to multipath propagation~\cite{heublein_feigl_crpa}, particularly under non-line-of-sight (NLoS) conditions. In contrast, ML-based approaches are data-driven and demonstrate robustness even in environments with strong multipath propagation. By exploiting the ability of wireless networks to capture extensive datasets, CIR samples can be associated with position labels to construct training databases~\cite{foliadis_garcia}. Nevertheless, fingerprinting-based methods~\cite{wang_jin_xia,foliadis_garcia_gallacher,lutakamale_myburgh,zhu_wang_sun_xu,kim_kim_moon,fu_liu_chen} present significant challenges, primarily due to the high cost of acquiring real-world training data. This problem is exacerbated in large-area applications where environmental dynamics necessitate frequent and costly retraining~\cite{stahlke2022transfer}. To address these limitations, this paper proposes a robust transfer learning technique capable of adapting to dynamic environmental changes, thereby reducing the dependency on extensive real-world data acquisition. In structured or semi-structured environments, an existing infrastructure for detecting and localizing changes is often already in place, thereby enhancing the effectiveness of selections.

\textit{Continual learning}~\cite{rusu_rabinowitz,zenke2017continual,rolnick_ahuja} refers to the sequential acquisition of novel tasks while mitigating \textit{catastrophic forgetting}, which occurs when models trained successively on multiple tasks lose the ability to perform well on previously learned tasks~\cite{kirkpatrick_rascanu,gaikwad_heublein}. Domain incremental learning (DIL) methods address this issue through approaches such as feature extraction with fine-tuning~\cite{li_hoiem} and by selectively slowing the learning of weights deemed critical for specific tasks~\cite{kirkpatrick_rascanu}. In the specific context of 5G indoor localization, transfer learning techniques~\cite{farzaneh_shahriar,foliadis_garcia_gallacher,etiabi_njima,dwedar_bayram,kim_kim_moon,kerdjidj_himeur_sohail} have been explored, whereas DIL approaches primarily focus on leveraging \textit{exemplars}, a small subset of previous data, to preserve task knowledge while adapting to new domains. These exemplars are carefully chosen based on criteria such as diversity, proximity to class prototypes, or informativeness and are typically employed for rehearsal, regularization, or alongside distillation methods~\cite{brignac_lobo,esaki_koide,mirza_masana,raichur_heublein}. To address the challenges posed by environmental dynamics in 5G indoor localization, we propose a similarity-aware approach that reduces the number of exemplars required for rapid adaptation.

\textbf{Contributions.} The primary objective of this work is to enhance model robustness for 5G indoor localization in dynamic and challenging environments characterized by varying propagation conditions, including diverse levels of channel complexity. To this end, we propose a DIL method that adapts to data from scenarios with environmental changes while mitigating catastrophic forgetting of knowledge from prior environments. Furthermore, we introduce a similarity-aware sampling technique that identifies and selects representative samples, referred to as \textit{exemplars}, which effectively capture the characteristics of the environment using KDTree for quick similarity computation. Consequently, we conduct a comparison of the following distance metrics: Euclidean, cosine similarity, Minkowski, Chebyshev, Canberra, Bray-Curtis, correlation, and Manhattan. We compare our approach with random, equally distributed, and error-dependent sample selection techniques. The proposed method effectively minimizes the number of selected samples, thereby reducing training time. Moreover, the model maintains a low localization error despite substantial environmental dynamics. Additionally, it prioritizes critical signal paths, ensuring high positioning accuracy even amid significant environmental changes.

\textbf{Outlook.} Sec.~\ref{label_related_work} offers a comprehensive review of existing literature on 5G indoor localization and DIL. Sec.~\ref{label_method} introduces the proposed DIL similarity-aware sampling technique. The dataset employed in this study is described in Sec.~\ref{label_experiments}, while Sec.~\ref{label_evaluation} presents a summary of the evaluation results.
\section{Related Work}
\label{label_related_work}

\subsection{5G Indoor Localization}
\label{label_rw_localization}

Stahlke et al.~\cite{stahlke2022transfer} investigated the generalizability of ML models, specifically CNNs, in realistic CIR setups with environmental variations, leveraging up to 25,000 adaptation samples. Their method achieves an error of $0.26\,m$ under identical training and testing conditions; however, the error increases to as much as $3.85\,m$ when the conditions are altered. Foliadis et al.~\cite{foliadis_garcia} assessed the effectiveness of early and late fusion techniques involving multiple BSs, with late fusion demonstrating superior positioning accuracy in dynamic scenarios. In a subsequent study \cite{foliadis_garcia_gallacher}, the authors proposed a multi-environment approach based on meta-learning, though it required up to 140,000 samples from target environments for effective adaptation. Vuckovic et al.~\cite{vuckovic_hosseini} introduced a fully connected autoencoder combined with Gaussian process regression, designed to operate with minimal labeled training data. However, this model exhibited adaptability only in scenarios sharing similar angle-delay profile characteristics. Wang et al.~\cite{wang_jin_xia} developed 5G1M, a framework utilizing a single 5G module integrated with a Siamese network incorporating Ghost modules and squeeze-and-excitation (SE) blocks, which mitigates dependency on extensive fingerprint databases. This model employs transfer learning to adapt to long-term environmental variations. Lutakamale et al.~\cite{lutakamale_myburgh} proposed a CNN architecture enhanced with SE blocks for node localization in LoRaWAN networks, demonstrating resilience even in noisy RSSI conditions. Chen et al.~\cite{chen_sarieddeen} examined the potential, challenges, and requirements of terahertz (THz) localization techniques. Whiton et al.~\cite{whiton_chen_tufvesson} utilized a software-defined radio (SDR) and a massive antenna array mounted on a ground vehicle to conduct experiments in urban NLoS scenarios. Farzaneh et al.~\cite{farzaneh_shahriar} evaluated the application of BiLSTM, CNN, and ResNet models for transfer learning in detecting distributed denial-of-service (DDoS) attacks within 5G networks. Dwedar et al.~\cite{dwedar_bayram} proposed a flexible ML model for anomaly detection in 5G networks, employing CNNs and transfer learning. Fu et al.~\cite{fu_liu_chen} introduced an auxiliary network to enhance WiFi-based positioning accuracy. Pan et al.~\cite{pan_wei_he} addressed catastrophic forgetting in DIL by employing EWC with $L_2$-regularization based on parameter importance. However, their approach relied solely on simulated signal parameters and necessitated a substantial fine-tuning dataset comprising 10,000 samples. Kabiri et al.~\cite{kabiri_cimarelli} and Ermolov et al.~\cite{ermolov_kadambi} combined 5G with inertial data.

\subsection{Domain Incremental Learning (DIL)}
\label{label_rw_dil}

Progressive neural networks (PNNs), proposed by Rusu et al.~\cite{rusu_rabinowitz}, are designed to prevent catastrophic forgetting while leveraging prior knowledge through lateral connections to previously learned features. Although PNNs offer robust memory retention, they suffer from scalability limitations due to architectural expansion with each new task, requiring additional lateral connections. Moreover, the method demands an understanding of task similarity and exhibits inflexibility in its network structure. Learning without forgetting (LwF)~\cite{li_hoiem} trains a network exclusively on data from the new task while preserving the original capabilities without revisiting previous task data. However, it assumes a relationship between the new and previously learned tasks. When this assumption is violated, particularly in cases where new tasks differ significantly, performance degradation may occur. Kirkpatrick et al.~\cite{kirkpatrick_rascanu} introduced elastic weight consolidation (EWC), which mitigates forgetting by selectively slowing learning on weights deemed important for previously learned tasks. Notably, EWC does not require access to prior task data, making it data-efficient. Synaptic intelligence (SI)~\cite{zenke2017continual} incorporates biologically inspired mechanisms, allowing each synapse to accumulate task-relevant information over time. This accumulated information facilitates the rapid storage of new memories while minimizing forgetting. However, SI stores importance measures for each parameter, leading to significant memory overhead as the number of tasks increases. Furthermore, its effectiveness may depend heavily on the similarity between tasks. Brignac et al.~\cite{brignac_lobo} focused on selecting the most informative samples for storage and determining the optimal number of samples using population strategies. In contrast, Esaki et al.~\cite{esaki_koide} proposed an approach specifically designed for the extreme case of one-shot DIL, where conventional methods perform inadequately. Additionally, Mirza et al.~\cite{mirza_masana} explored the storage of first- and second-order statistical parameters to address variations in weather conditions for autonomous driving systems.

In this work, we leverage Finetune, PNN, EWC, LwF, and SI as state-of-the-art DIL methods, while extending LwF and EWC with similarity-aware sampling strategies. By enabling EWC to access previous task data for similarity computation, we aim to further enhance its performance in scenarios involving diverse task distributions.

\begin{figure*}[!t]
    \centering
    \includegraphics[width=1.0\linewidth]{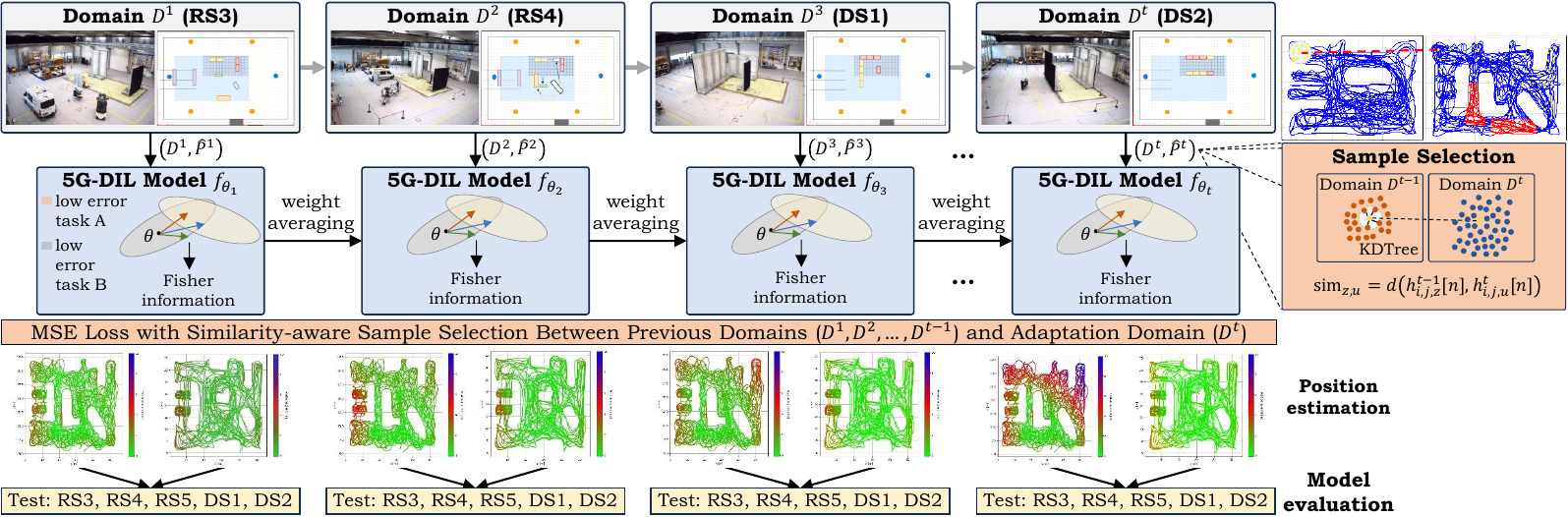}
    \caption{\textbf{Overview of our 5G-DIL method.} The model is trained using CIR data $D^t$ and the corresponding position labels $P^t$ for each task $t$ within the 5G dataset~\cite{stahlke2022transfer}. To incrementally learn across domains without incurring catastrophic forgetting, the model updates its parameters by averaging the weights obtained from previous tasks $\{t-1, t-2, \ldots, 1\}$. Our similarity-aware sample selection strategy is employed to choose a limited number of samples ($N \leq 200$) from the adaptation domain, enabling efficient fine-tuning. After the completion of each task, the model is evaluated across all previously encountered domains.}
    \label{figure_method_overview}
\end{figure*}

\section{Methodology}
\label{label_method}

Initially, we present the notation for DIL. Subsequently, we provide an overview of our 5G-DIL method and elaborate on the similarity-aware sample selection technique.

\textbf{Notation.} In 5G indoor localization, the CIR serves as a critical parameter for characterizing the propagation environment between a transmitter and a receiver. It encapsulates the multipath effects of the transmitted signal as it undergoes reflections from various surfaces before arriving at the receiver~\cite{stahlke2022transfer}. The CIR is commonly expressed as a complex-valued function $h(t)$, where $t$ represents time. In its discrete-time formulation, the CIR is modeled as a sequence $h[n]$, with $n$ indexing the discrete time samples. For systems employing multiple antennas, the CIR can be extended to $h_{i,j}[n]$, where $i$ and $j$ correspond to the indices of the transmitting and receiving antennas, respectively. The primary objective is to estimate the position of a user or device, denoted by $\hat{\textbf{p}}_{i,j} = (\hat{x},\hat{y}) \in \mathbb{R}^2$, representing its coordinates in a two-dimensional space. The reference positions are denoted as $\textbf{p}_{i,j} = (x,y) \in \mathbb{R}^2$~\cite{foliadis_garcia}. In incremental learning scenarios, a sequence of tasks is learned one at a time within their own training set, which can be defined as the sequence of $k$ tasks $\mathcal{T} = [(D^1, {P}^1), (D^2, {P}^2), \ldots, (D^k, {P}^k)]$, where the set of reference positions ${P}^t = \{{\mathbf{p}}_{i,j,1}^t, {\mathbf{p}}_{i,j,2}^t, \ldots, {\mathbf{p}}_{i,j,k^t}^t\}$ represents each task $t$, learned with training data $D^t$~\cite{mirza_masana}. The training set consists of the input features $h_{i,j}[n]$ (i.e., CIRs) and corresponding ground truth labels ${\textbf{p}}_{i,j}$ (i.e., positions). Therefore, we have
\begin{equation}
    \resizebox{0.91\linewidth}{!}{$
    \displaystyle
    (D^t, {P}^t) = \{(h_{i,j,1}^t[n], {\textbf{p}}_{i,j,1}^t), (h_{i,j,2}^t[n], {\textbf{p}}_{i,j,2}^t), \ldots, (h_{i,j,k^t}^t[n], {\textbf{p}}_{i,j,k^t}^t)\}
    $}
\end{equation}
all available pairs of input and labels for a given task, where all ${\textbf{p}}_{i,j}^t \in {P}^t$~\cite{mirza_masana}. In contrast to \textit{non-exemplar}-based DIL, which does not store any samples from previous domains and instead relies on regularization techniques to retain knowledge from prior tasks, \textit{exemplar}-based DIL carefully selects and stores a subset of exemplars. Let $f_{\theta_{t}}$ represent the neural network parameterized by $\theta_t$ after task $t$, which learns a mapping from the input $h_{i,j}^t[n]$ to the corresponding output $\textbf{p}_{i,j}^t$. Denote $\epsilon$ as the exemplar set, a small subset of the adaptation domain data from tasks $t$, consisting of $N$ samples. The goal is to incrementally train $f_{\theta_{t}}$ on new domains while minimizing the loss over all past and current tasks~\cite{wang_he_dong,shi_wang}. Our objective is to select the most representative exemplars $\epsilon$ from the adaptation domain while minimizing the number of samples $N$ to facilitate efficient model training by reducing the domain shift between domains~\cite{wang_gong,ott_acmmm}.

\textbf{Method Overview.} Figure~\ref{figure_method_overview} provides a comprehensive overview of the proposed 5G-DIL method. For each task $t$, the model processes the CIR measurements $D^t$ along with the corresponding reference positions ${P}^t$. The model $f_{\theta_t}$ is optimized to predict the trajectory of the user equipment by minimizing the mean squared error (MSE) loss. To address novel domains affected by environmental changes, the model adaptively fine-tunes its parameters after each task while mitigating catastrophic forgetting by averaging the current and previous model weights. Given the resource-intensive nature of fine-tuning on the entire scenario, a more efficient approach is adopted. Specifically, the model is trained only on the modified regions of the new scenario. Additionally, a significantly smaller number of samples, 
$N$, with values of 0, 50, 100, and 200, are selected from the non-modified regions that are similar to the previous scenario. These selected samples, referred to as exemplars, aim to identify the most representative samples by computing errors and assessing similarities between the previous and adaptation domains. Details of this process are discussed in the subsequent section.

\begin{figure*}[!t]\captionsetup[subfigure]{font=scriptsize}
    \centering
	\begin{minipage}[t]{0.205\linewidth}
        \centering
        \includegraphics[width=1.0\linewidth]{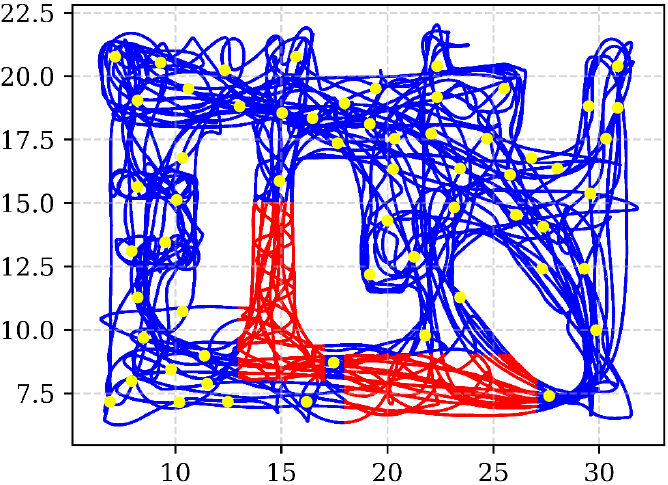}
        \subcaption{Randomly distributed (RD).}
        \label{figure_sample_selection1}
    \end{minipage}
    \hfill
	\begin{minipage}[t]{0.205\linewidth}
        \centering
        \includegraphics[width=1.0\linewidth]{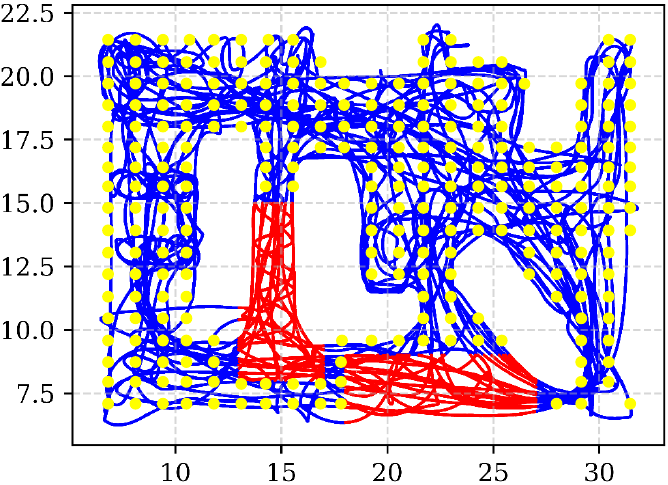}
        \subcaption{Equally distributed (ED).}
        \label{figure_sample_selection2}
    \end{minipage}
    \hfill
	\begin{minipage}[t]{0.145\linewidth}
        \centering
        \includegraphics[width=1.0\linewidth]{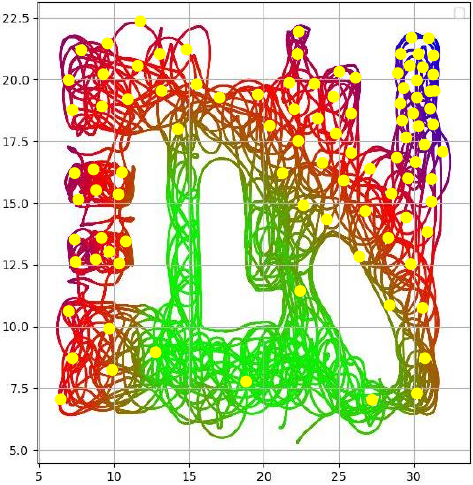}
        \subcaption{Error-dependent.}
        \label{figure_sample_selection3}
    \end{minipage}
    \hfill
	\begin{minipage}[t]{0.41\linewidth}
        \centering
        \includegraphics[width=1.0\linewidth]{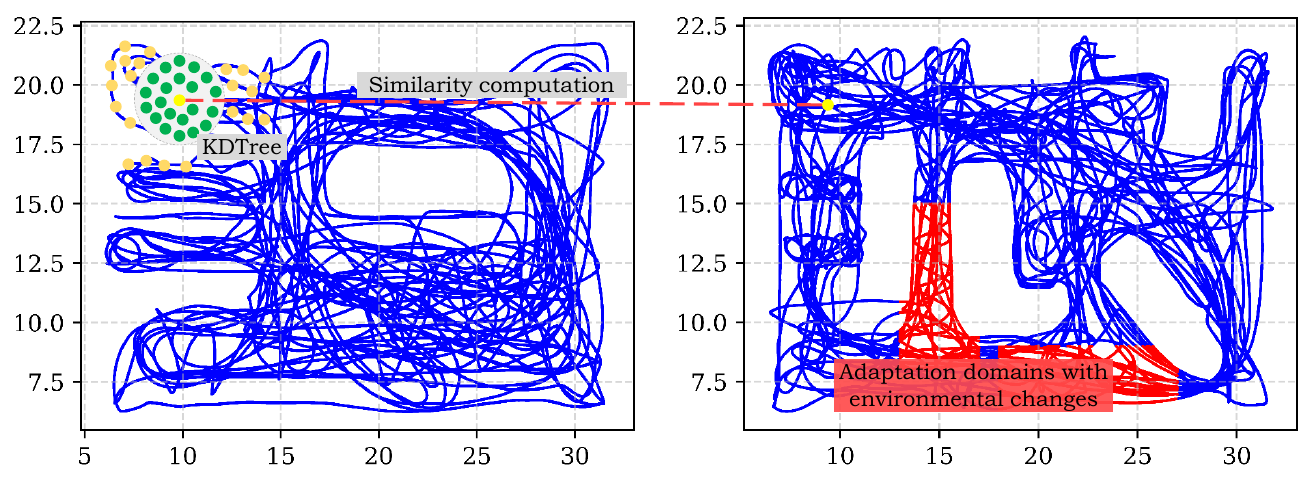}
        \subcaption{Similarity-based.}
        \label{figure_sample_selection4}
    \end{minipage}
    \caption{Sample selection for task $D^t$ (\textit{red}: environmental changes, \textit{blue}: static environment, \textit{yellow} dots: selected samples).}
    \label{figure_sample_selection}
\end{figure*}

\textbf{Sample Selection.} Figure~\ref{figure_sample_selection} illustrates the various sample selection techniques. The objective is to select $N$ samples from the adaptation domain of task $t$ that optimally represent the domain characteristics and capture domain shifts, thereby minimizing positioning error while ensuring efficient fine-tuning runtime. The baseline approach involves \textit{randomly distributed} (RD) selecting $N$ samples from the domain dataset based on a uniform distribution (see Figure~\ref{figure_sample_selection1}). To achieve a better spatial representation across the entire coordinate system of the domain, we consider an \textit{equally distributed} (ED) sample selection (uniform) strategy (see Figure~\ref{figure_sample_selection2}). However, this approach is suboptimal due to the presence of multipath effects in certain regions, which can introduce data ambiguities and consequently impair position estimation accuracy~\cite{nikonowicz_mahmood}. An enhanced strategy involves selecting samples with either the \textit{highest or lowest estimation errors}. High-error samples typically indicate regions where the model's understanding is insufficient. Training on these samples allows the model to learn new features, thereby addressing its knowledge gaps. Additionally, this approach enhances the model's robustness to challenging or ambiguous cases, improves generalization across diverse data distributions, and ensures focus on the most informative data points, thereby potentially accelerating the learning process and improving convergence. Conversely, revisiting low-error samples helps reinforce the model’s existing knowledge and prevents the forgetting of previously learned information. Moreover, incorporating such samples provides a stable foundation during training and effectively mitigates catastrophic forgetting~\cite{kalb_mauthe_beyerer,zajac_tuytelaars}. The CIR data from domain $t$ is denoted as $h_{i,j,l}^t[n]$ for sample $l$. The estimation error between the predicted and reference positions is defined as $e_{i,j,l}^t = \text{Euclidean}\big(\hat{\textbf{p}}_{i,j,l}^t, \textbf{p}_{i,j,l}^t\big)$ in 2D space, where $l \in \{1, 2, \dots, k^t\}$, and $k^t$ represents the total number of samples in task $t$. Subsequently, $N$ samples are selected based on either the highest or lowest position errors $e$. Figure~\ref{figure_sample_selection3} provides a visualization of this error-dependent selection strategy. Our approach calculates the \textit{similarity} between all samples in non-changing environments from the previous domain $D^{t-1}$ to the adaptation domain $D^{t}$, as defined by
\begin{equation}
    \text{sim}_{z,u} = d \big( h_{i,j,z}^{t-1}[n] - h_{i,j,u}^{t}[n] \big), \, \forall \, i, j,
\end{equation}
where $d$ denotes a distance metric, and let $u \in \{1, \ldots, k^{t}\}$ represent all samples from the adaptation domain. For each sample $u$, we identify the element $z \in \{1, \ldots, k^{t-1}\}$ that minimizes the distance to $u$. This process leverages the \texttt{KDTree} implementation from \textit{scipy.spatial.KDTree}~\cite{scipy_kdtree} to efficiently perform nearest-neighbor searches. The selection process is illustrated in Figure~\ref{figure_sample_selection4}. We evaluate various metrics, including Euclidean, cosine similarity, Minkowski, Chebyshev, Canberra, Bray-Curtis, correlation, and Manhattan distances. Subsequently, we select the top $N$ samples from the adaptation domain $D^t$ that exhibit the highest similarity, thereby reinforcing the model's existing knowledge and mitigating the risk of forgetting previously acquired information.
\section{Experiments}
\label{label_experiments}

\begin{table}[!t]
\begin{center}
\setlength{\tabcolsep}{3.1pt}
    \caption{Overview of (initial) training (full dataset), adaptation (subset with $N$ exemplars), and test datasets. $\rightarrow$ defines the use of weights from previous adaptations.}
    \label{table_dataset_overview}
    \begin{tabular}{ p{0.5cm} | p{0.5cm} | p{0.5cm} }
    \multicolumn{1}{c|}{\textbf{Training}} & \multicolumn{1}{c|}{\textbf{$\rightarrow$ Adaptation Set}}& \multicolumn{1}{c}{\textbf{Test}} \\ \hline
    \multicolumn{1}{c|}{RS3} & \multicolumn{1}{l|}{} & \multicolumn{1}{l}{RS3, RS4, RS5, DS1, DS2} \\
    \multicolumn{1}{c|}{RS3} & \multicolumn{1}{l|}{$\rightarrow$ RS4} & \multicolumn{1}{l}{RS3, RS4, RS5, DS1, DS2} \\
    \multicolumn{1}{c|}{RS3} & \multicolumn{1}{l|}{$\rightarrow$ RS4 $\rightarrow$ RS5} & \multicolumn{1}{l}{RS3, RS4, RS5, DS1, DS2} \\
    \multicolumn{1}{c|}{RS3} & \multicolumn{1}{l|}{$\rightarrow$ RS4 $\rightarrow$ RS5 $\rightarrow$ DS1} & \multicolumn{1}{l}{RS3, RS4, RS5, DS1, DS2} \\
    \multicolumn{1}{c|}{RS3} & \multicolumn{1}{l|}{$\rightarrow$ RS4 $\rightarrow$ RS5 $\rightarrow$ DS1 $\rightarrow$ DS2} & \multicolumn{1}{l}{RS3, RS4, RS5, DS1, DS2} \\
    \hline
    \multicolumn{1}{c|}{RS3} & \multicolumn{1}{l|}{$\rightarrow$ DS1} & \multicolumn{1}{l}{RS3, DS1} \\
    \end{tabular}
\end{center}
\end{table}

\begin{table*}[!t]
\begin{center}
\setlength{\tabcolsep}{2.0pt}
    \caption{Evaluation results of all DIL methods for RS3 (full set) $\rightarrow$ RS4 $\rightarrow$ RS5 $\rightarrow$ DS1 $\rightarrow$ DS2. Number of exemplars: $N=0$ (row 1), $N=50$ (row 2), $N=100$ (row 3), and $N=200$ (row 4). \textbf{Bold}: best results for each $N$. \underline{Underline}: overall.}
    \label{table_dataset_results}
    \scriptsize \begin{tabular}{ p{0.5cm} | p{0.5cm} | p{0.5cm} | p{0.5cm} | p{0.5cm} | p{0.5cm} | p{0.5cm} | p{0.5cm} | p{0.5cm} | p{0.5cm} | p{0.5cm} | p{0.5cm} | p{0.5cm} | p{0.5cm} | p{0.5cm} | p{0.5cm} | p{0.5cm} | p{0.5cm} | p{0.5cm} | p{0.5cm} | p{0.5cm} | p{0.5cm} | p{0.5cm} | p{0.5cm} | p{0.5cm} | p{0.5cm} }
    \multicolumn{1}{c|}{} & \multicolumn{5}{c|}{\textbf{RS3}} & \multicolumn{5}{c|}{\textbf{RS4}} & \multicolumn{5}{c|}{\textbf{RS5}} & \multicolumn{5}{c|}{\textbf{DS1}} & \multicolumn{5}{c}{\textbf{DS2}} \\
    \multicolumn{1}{c|}{} & \multicolumn{1}{c}{\textbf{EWC}} & \multicolumn{1}{c}{\textbf{FT}} & \multicolumn{1}{c}{\textbf{LwF}} & \multicolumn{1}{c}{\textbf{SI}} & \multicolumn{1}{c|}{\textbf{PNN}} & \multicolumn{1}{c}{\textbf{EWC}} & \multicolumn{1}{c}{\textbf{FT}} & \multicolumn{1}{c}{\textbf{LwF}} & \multicolumn{1}{c}{\textbf{SI}} & \multicolumn{1}{c|}{\textbf{PNN}} & \multicolumn{1}{c}{\textbf{EWC}} & \multicolumn{1}{c}{\textbf{FT}} & \multicolumn{1}{c}{\textbf{LwF}} & \multicolumn{1}{c}{\textbf{SI}} & \multicolumn{1}{c|}{\textbf{PNN}} & \multicolumn{1}{c}{\textbf{EWC}} & \multicolumn{1}{c}{\textbf{FT}} & \multicolumn{1}{c}{\textbf{LwF}} & \multicolumn{1}{c}{\textbf{SI}} & \multicolumn{1}{c|}{\textbf{PNN}} & \multicolumn{1}{c}{\textbf{EWC}} & \multicolumn{1}{c}{\textbf{FT}} & \multicolumn{1}{c}{\textbf{LwF}} & \multicolumn{1}{c}{\textbf{SI}} & \multicolumn{1}{c}{\textbf{PNN}} \\ \hline
    \multicolumn{1}{c|}{\textbf{RS3}} & \multicolumn{1}{r}{0.235} & \multicolumn{1}{r}{0.235} & \multicolumn{1}{r}{0.235} & \multicolumn{1}{r}{\underline{\textbf{0.160}}} & \multicolumn{1}{r|}{0.236} & \multicolumn{1}{r}{0.325} & \multicolumn{1}{r}{0.325} & \multicolumn{1}{r}{0.325} & \multicolumn{1}{r}{\underline{\textbf{0.308}}} & \multicolumn{1}{r|}{0.325} & \multicolumn{1}{r}{\underline{\textbf{0.291}}} & \multicolumn{1}{r}{\underline{\textbf{0.291}}} & \multicolumn{1}{r}{\underline{\textbf{0.291}}} & \multicolumn{1}{r}{0.317} & \multicolumn{1}{r|}{\underline{\textbf{0.291}}} & \multicolumn{1}{r}{0.634} & \multicolumn{1}{r}{0.634} & \multicolumn{1}{r}{0.634} & \multicolumn{1}{r}{\underline{\textbf{0.618}}} & \multicolumn{1}{r|}{0.634} & \multicolumn{1}{r}{\underline{\textbf{0.632}}} & \multicolumn{1}{r}{\underline{\textbf{0.632}}} & \multicolumn{1}{r}{\underline{\textbf{0.632}}} & \multicolumn{1}{r}{0.649} & \multicolumn{1}{r}{\underline{\textbf{0.632}}} \\ \hline
    \multirow{4}{*}{\textbf{RS4}} & \multicolumn{1}{r}{\textbf{0.431}} & \multicolumn{1}{r}{1.217} & \multicolumn{1}{r}{0.787} & \multicolumn{1}{r}{1.147} & \multicolumn{1}{r|}{0.653} & \multicolumn{1}{r}{\textbf{0.486}} & \multicolumn{1}{r}{1.347} & \multicolumn{1}{r}{0.958} & \multicolumn{1}{r}{1.166} & \multicolumn{1}{r|}{0.856} & \multicolumn{1}{r}{\textbf{0.460}} & \multicolumn{1}{r}{1.075} & \multicolumn{1}{r}{0.775} & \multicolumn{1}{r}{1.079} & \multicolumn{1}{r|}{0.650} & \multicolumn{1}{r}{\textbf{0.646}} & \multicolumn{1}{r}{1.311} & \multicolumn{1}{r}{1.003} & \multicolumn{1}{r}{1.204} & \multicolumn{1}{r|}{0.857} & \multicolumn{1}{r}{\textbf{0.636}} & \multicolumn{1}{r}{1.375} & \multicolumn{1}{r}{1.033} & \multicolumn{1}{r}{1.314} & \multicolumn{1}{r}{0.970} \\
    \multicolumn{1}{c|}{} & \multicolumn{1}{r}{\textbf{0.385}} & \multicolumn{1}{r}{0.587} & \multicolumn{1}{r}{0.493} & \multicolumn{1}{r}{0.813} & \multicolumn{1}{r|}{0.613} & \multicolumn{1}{r}{\textbf{0.442}} & \multicolumn{1}{r}{0.652} & \multicolumn{1}{r}{0.650} & \multicolumn{1}{r}{0.873} & \multicolumn{1}{r|}{0.645} & \multicolumn{1}{r}{\textbf{0.420}} & \multicolumn{1}{r}{0.443} & \multicolumn{1}{r}{0.540} & \multicolumn{1}{r}{0.708} & \multicolumn{1}{r|}{0.779} & \multicolumn{1}{r}{\textbf{0.616}} & \multicolumn{1}{r}{0.808} & \multicolumn{1}{r}{0.803} & \multicolumn{1}{r}{0.946} & \multicolumn{1}{r|}{0.711} & \multicolumn{1}{r}{\textbf{0.596}} & \multicolumn{1}{r}{0.756} & \multicolumn{1}{r}{0.805} & \multicolumn{1}{r}{0.884} & \multicolumn{1}{r}{0.710} \\
    \multicolumn{1}{c|}{} & \multicolumn{1}{r}{0.364} & \multicolumn{1}{r}{0.498} & \multicolumn{1}{r}{\textbf{0.323}} & \multicolumn{1}{r}{0.643} & \multicolumn{1}{r|}{0.682} & \multicolumn{1}{r}{\textbf{0.421}} & \multicolumn{1}{r}{0.513} & \multicolumn{1}{r}{0.443} & \multicolumn{1}{r}{0.680} & \multicolumn{1}{r|}{0.834} & \multicolumn{1}{r}{0.395} & \multicolumn{1}{r}{\textbf{0.347}} & \multicolumn{1}{r}{0.373} & \multicolumn{1}{r}{0.545} & \multicolumn{1}{r|}{0.663} & \multicolumn{1}{r}{\textbf{0.590}} & \multicolumn{1}{r}{0.756} & \multicolumn{1}{r}{0.647} & \multicolumn{1}{r}{0.819} & \multicolumn{1}{r|}{0.965} & \multicolumn{1}{r}{\textbf{0.588}} & \multicolumn{1}{r}{0.727} & \multicolumn{1}{r}{0.635} & \multicolumn{1}{r}{0.745} & \multicolumn{1}{r}{1.064} \\
    \multicolumn{1}{c|}{} & \multicolumn{1}{r}{0.331} & \multicolumn{1}{r}{0.358} & \multicolumn{1}{r}{\underline{\textbf{0.296}}} & \multicolumn{1}{r}{0.479} & \multicolumn{1}{r|}{0.418} & \multicolumn{1}{r}{\underline{\textbf{0.385}}} & \multicolumn{1}{r}{0.392} & \multicolumn{1}{r}{0.404} & \multicolumn{1}{r}{0.499} & \multicolumn{1}{r|}{0.645} & \multicolumn{1}{r}{0.340} & \multicolumn{1}{r}{\underline{\textbf{0.281}}} & \multicolumn{1}{r}{0.302} & \multicolumn{1}{r}{0.438} & \multicolumn{1}{r|}{0.480} & \multicolumn{1}{r}{\underline{\textbf{0.589}}} & \multicolumn{1}{r}{0.630} & \multicolumn{1}{r}{0.606} & \multicolumn{1}{r}{0.667} & \multicolumn{1}{r|}{0.955} & \multicolumn{1}{r}{\underline{\textbf{0.576}}} & \multicolumn{1}{r}{0.607} & \multicolumn{1}{r}{0.618} & \multicolumn{1}{r}{0.678} & \multicolumn{1}{r}{0.710} \\ \hline
    \multirow{4}{*}{\textbf{RS5}} & \multicolumn{1}{r}{\textbf{0.488}} & \multicolumn{1}{r}{1.311} & \multicolumn{1}{r}{1.101} & \multicolumn{1}{r}{1.264} & \multicolumn{1}{r|}{0.925} & \multicolumn{1}{r}{\textbf{0.534}} & \multicolumn{1}{r}{1.418} & \multicolumn{1}{r}{1.241} & \multicolumn{1}{r}{1.325} & \multicolumn{1}{r|}{1.021} & \multicolumn{1}{r}{\textbf{0.466}} & \multicolumn{1}{r}{1.089} & \multicolumn{1}{r}{0.962} & \multicolumn{1}{r}{1.140} & \multicolumn{1}{r|}{0.856} & \multicolumn{1}{r}{\textbf{0.716}} & \multicolumn{1}{r}{1.357} & \multicolumn{1}{r}{1.106} & \multicolumn{1}{r}{1.306} & \multicolumn{1}{r|}{0.921} & \multicolumn{1}{r}{\textbf{0.699}} & \multicolumn{1}{r}{1.437} & \multicolumn{1}{r}{1.228} & \multicolumn{1}{r}{1.440} & \multicolumn{1}{r}{1.004} \\
    \multicolumn{1}{c|}{} & \multicolumn{1}{r}{\textbf{0.413}} & \multicolumn{1}{r}{0.649} & \multicolumn{1}{r}{0.591} & \multicolumn{1}{r}{0.690} & \multicolumn{1}{r|}{0.540} & \multicolumn{1}{r}{\textbf{0.454}} & \multicolumn{1}{r}{0.789} & \multicolumn{1}{r}{0.714} & \multicolumn{1}{r}{0.692} & \multicolumn{1}{r|}{0.678} & \multicolumn{1}{r}{\textbf{0.392}} & \multicolumn{1}{r}{0.478} & \multicolumn{1}{r}{0.560} & \multicolumn{1}{r}{0.554} & \multicolumn{1}{r|}{0.520} & \multicolumn{1}{r}{\textbf{0.657}} & \multicolumn{1}{r}{0.767} & \multicolumn{1}{r}{0.853} & \multicolumn{1}{r}{0.892} & \multicolumn{1}{r|}{0.801} & \multicolumn{1}{r}{\textbf{0.645}} & \multicolumn{1}{r}{0.749} & \multicolumn{1}{r}{0.829} & \multicolumn{1}{r}{0.819} & \multicolumn{1}{r}{0.816} \\
    \multicolumn{1}{c|}{} & \multicolumn{1}{r}{\textbf{0.368}} & \multicolumn{1}{r}{0.430} & \multicolumn{1}{r}{0.439} & \multicolumn{1}{r}{0.583} & \multicolumn{1}{r|}{0.687} & \multicolumn{1}{r}{\textbf{0.417}} & \multicolumn{1}{r}{0.472} & \multicolumn{1}{r}{0.537} & \multicolumn{1}{r}{0.566} & \multicolumn{1}{r|}{0.783} & \multicolumn{1}{r}{0.354} & \multicolumn{1}{r}{\textbf{0.336}} & \multicolumn{1}{r}{0.480} & \multicolumn{1}{r}{0.479} & \multicolumn{1}{r|}{0.573} & \multicolumn{1}{r}{\textbf{0.604}} & \multicolumn{1}{r}{0.660} & \multicolumn{1}{r}{0.693} & \multicolumn{1}{r}{0.883} & \multicolumn{1}{r|}{0.961} & \multicolumn{1}{r}{\textbf{0.611}} & \multicolumn{1}{r}{0.640} & \multicolumn{1}{r}{0.675} & \multicolumn{1}{r}{0.751} & \multicolumn{1}{r}{0.966}  \\
    \multicolumn{1}{c|}{} & \multicolumn{1}{r}{\underline{\textbf{0.314}}} & \multicolumn{1}{r}{0.319} & \multicolumn{1}{r}{0.357} & \multicolumn{1}{r}{0.455} & \multicolumn{1}{r|}{0.540} & \multicolumn{1}{r}{0.365} & \multicolumn{1}{r}{\underline{\textbf{0.358}}} & \multicolumn{1}{r}{0.436} & \multicolumn{1}{r}{0.485} & \multicolumn{1}{r|}{0.678} & \multicolumn{1}{r}{0.298} & \multicolumn{1}{r}{\underline{\textbf{0.261}}} & \multicolumn{1}{r}{0.321} & \multicolumn{1}{r}{0.382} & \multicolumn{1}{r|}{0.520} & \multicolumn{1}{r}{\underline{\textbf{0.573}}} & \multicolumn{1}{r}{0.580} & \multicolumn{1}{r}{0.616} & \multicolumn{1}{r}{0.788} & \multicolumn{1}{r|}{0.801} & \multicolumn{1}{r}{0.594} & \multicolumn{1}{r}{\underline{\textbf{0.585}}} & \multicolumn{1}{r}{0.626} & \multicolumn{1}{r}{0.695} & \multicolumn{1}{r}{0.816} \\ \hline
    \multirow{4}{*}{\textbf{DS1}} & \multicolumn{1}{r}{\textbf{0.693}} & \multicolumn{1}{r}{1.318} & \multicolumn{1}{r}{0.839} & \multicolumn{1}{r}{1.000} & \multicolumn{1}{r|}{1.493} & \multicolumn{1}{r}{\textbf{0.760}} & \multicolumn{1}{r}{1.263} & \multicolumn{1}{r}{1.043} & \multicolumn{1}{r}{1.017} & \multicolumn{1}{r|}{1.276} & \multicolumn{1}{r}{\textbf{0.570}} & \multicolumn{1}{r}{1.059} & \multicolumn{1}{r}{0.827} & \multicolumn{1}{r}{0.848} & \multicolumn{1}{r|}{1.113} & \multicolumn{1}{r}{0.384} & \multicolumn{1}{r}{0.416} & \multicolumn{1}{r}{0.887} & \multicolumn{1}{r}{\textbf{0.359}} & \multicolumn{1}{r|}{0.667} & \multicolumn{1}{r}{0.641} & \multicolumn{1}{r}{0.770} & \multicolumn{1}{r}{0.972} & \multicolumn{1}{r}{0.748} & \multicolumn{1}{r}{\textbf{0.628}} \\
    \multicolumn{1}{c|}{} & \multicolumn{1}{r}{0.657} & \multicolumn{1}{r}{0.832} & \multicolumn{1}{r}{\textbf{0.581}} & \multicolumn{1}{r}{0.875} & \multicolumn{1}{r|}{1.231} & \multicolumn{1}{r}{\textbf{0.703}} & \multicolumn{1}{r}{1.073} & \multicolumn{1}{r}{0.730} & \multicolumn{1}{r}{0.887} & \multicolumn{1}{r|}{1.119} & \multicolumn{1}{r}{0.542} & \multicolumn{1}{r}{0.794} & \multicolumn{1}{r}{\textbf{0.525}} & \multicolumn{1}{r}{0.712} & \multicolumn{1}{r|}{1.120} & \multicolumn{1}{r}{0.376} & \multicolumn{1}{r}{\textbf{0.299}} & \multicolumn{1}{r}{0.676} & \multicolumn{1}{r}{0.370} & \multicolumn{1}{r|}{0.600} & \multicolumn{1}{r}{0.643} & \multicolumn{1}{r}{0.702} & \multicolumn{1}{r}{0.719} & \multicolumn{1}{r}{0.724} & \multicolumn{1}{r}{\textbf{0.611}} \\
    \multicolumn{1}{c|}{} & \multicolumn{1}{r}{0.635} & \multicolumn{1}{r}{0.723} & \multicolumn{1}{r}{\textbf{0.493}} & \multicolumn{1}{r}{0.851} & \multicolumn{1}{r|}{1.296} & \multicolumn{1}{r}{0.683} & \multicolumn{1}{r}{0.856} & \multicolumn{1}{r}{\textbf{0.615}} & \multicolumn{1}{r}{0.724} & \multicolumn{1}{r|}{1.207} & \multicolumn{1}{r}{0.544} & \multicolumn{1}{r}{0.774} & \multicolumn{1}{r}{\textbf{0.483}} & \multicolumn{1}{r}{0.634} & \multicolumn{1}{r|}{1.041} & \multicolumn{1}{r}{0.356} & \multicolumn{1}{r}{\textbf{0.266}} & \multicolumn{1}{r}{0.608} & \multicolumn{1}{r}{0.345} & \multicolumn{1}{r|}{0.512} & \multicolumn{1}{r}{0.653} & \multicolumn{1}{r}{\textbf{0.578}} & \multicolumn{1}{r}{0.643} & \multicolumn{1}{r}{0.715} & \multicolumn{1}{r}{0.628} \\
    \multicolumn{1}{c|}{} & \multicolumn{1}{r}{0.649} & \multicolumn{1}{r}{0.937} & \multicolumn{1}{r}{\underline{\textbf{0.448}}} & \multicolumn{1}{r}{0.673} & \multicolumn{1}{r|}{1.128} & \multicolumn{1}{r}{0.693} & \multicolumn{1}{r}{0.961} & \multicolumn{1}{r}{\underline{\textbf{0.550}}} & \multicolumn{1}{r}{0.695} & \multicolumn{1}{r|}{1.119} & \multicolumn{1}{r}{0.554} & \multicolumn{1}{r}{0.835} & \multicolumn{1}{r}{\underline{\textbf{0.412}}} & \multicolumn{1}{r}{0.666} & \multicolumn{1}{r|}{1.003} & \multicolumn{1}{r}{0.361} & \multicolumn{1}{r}{0.372} & \multicolumn{1}{r}{0.539} & \multicolumn{1}{r}{\underline{\textbf{0.307}}} & \multicolumn{1}{r|}{0.505} & \multicolumn{1}{r}{0.626} & \multicolumn{1}{r}{\underline{\textbf{0.553}}} & \multicolumn{1}{r}{0.608} & \multicolumn{1}{r}{0.599} & \multicolumn{1}{r}{0.611} \\ \hline
    \multirow{4}{*}{\textbf{DS2}} & \multicolumn{1}{r}{\textbf{0.760}} & \multicolumn{1}{r}{1.143} & \multicolumn{1}{r}{0.855} & \multicolumn{1}{r}{1.264} & \multicolumn{1}{r|}{1.551} & \multicolumn{1}{r}{\textbf{0.791}} & \multicolumn{1}{r}{1.118} & \multicolumn{1}{r}{1.069} & \multicolumn{1}{r}{1.261} & \multicolumn{1}{r|}{1.345} & \multicolumn{1}{r}{\textbf{0.586}} & \multicolumn{1}{r}{0.875} & \multicolumn{1}{r}{0.804} & \multicolumn{1}{r}{1.066} & \multicolumn{1}{r|}{1.230} & \multicolumn{1}{r}{\textbf{0.509}} & \multicolumn{1}{r}{0.761} & \multicolumn{1}{r}{0.907} & \multicolumn{1}{r}{0.689} & \multicolumn{1}{r|}{0.737} & \multicolumn{1}{r}{0.481} & \multicolumn{1}{r}{0.549} & \multicolumn{1}{r}{0.929} & \multicolumn{1}{r}{\textbf{0.455}} & \multicolumn{1}{r}{0.734} \\
    \multicolumn{1}{c|}{} & \multicolumn{1}{r}{0.719} & \multicolumn{1}{r}{1.138} & \multicolumn{1}{r}{\textbf{0.666}} & \multicolumn{1}{r}{1.154} & \multicolumn{1}{r|}{1.329} & \multicolumn{1}{r}{\textbf{0.751}} & \multicolumn{1}{r}{1.106} & \multicolumn{1}{r}{0.891} & \multicolumn{1}{r}{1.118} & \multicolumn{1}{r|}{1.228} & \multicolumn{1}{r}{0.566} & \multicolumn{1}{r}{0.984} & \multicolumn{1}{r}{\textbf{0.558}} & \multicolumn{1}{r}{0.838} & \multicolumn{1}{r|}{1.112} & \multicolumn{1}{r}{\underline{\textbf{0.465}}} & \multicolumn{1}{r}{0.822} & \multicolumn{1}{r}{0.667} & \multicolumn{1}{r}{0.682} & \multicolumn{1}{r|}{0.708} & \multicolumn{1}{r}{0.473} & \multicolumn{1}{r}{\textbf{0.365}} & \multicolumn{1}{r}{0.661} & \multicolumn{1}{r}{0.435} & \multicolumn{1}{r}{0.698} \\
    \multicolumn{1}{c|}{} & \multicolumn{1}{r}{0.731} & \multicolumn{1}{r}{0.983} & \multicolumn{1}{r}{\underline{\textbf{0.538}}} & \multicolumn{1}{r}{0.930} & \multicolumn{1}{r|}{1.322} & \multicolumn{1}{r}{0.764} & \multicolumn{1}{r}{0.941} & \multicolumn{1}{r}{\underline{\textbf{0.685}}} & \multicolumn{1}{r}{0.903} & \multicolumn{1}{r|}{1.230} & \multicolumn{1}{r}{0.624} & \multicolumn{1}{r}{0.856} & \multicolumn{1}{r}{\underline{\textbf{0.468}}} & \multicolumn{1}{r}{0.791} & \multicolumn{1}{r|}{1.054} & \multicolumn{1}{r}{\textbf{0.468}} & \multicolumn{1}{r}{0.625} & \multicolumn{1}{r}{0.616} & \multicolumn{1}{r}{0.594} & \multicolumn{1}{r|}{0.684} & \multicolumn{1}{r}{0.456} & \multicolumn{1}{r}{0.339} & \multicolumn{1}{r}{0.591} & \multicolumn{1}{r}{\textbf{0.358}} & \multicolumn{1}{r}{0.684} \\
    \multicolumn{1}{c|}{} & \multicolumn{1}{r}{0.672} & \multicolumn{1}{r}{0.886} & \multicolumn{1}{r}{\textbf{0.594}} & \multicolumn{1}{r}{0.806} & \multicolumn{1}{r|}{1.229} & \multicolumn{1}{r}{0.709} & \multicolumn{1}{r}{0.844} & \multicolumn{1}{r}{\textbf{0.707}} & \multicolumn{1}{r}{0.806} & \multicolumn{1}{r|}{1.237} & \multicolumn{1}{r}{0.565} & \multicolumn{1}{r}{0.744} & \multicolumn{1}{r}{\textbf{0.531}} & \multicolumn{1}{r}{0.922} & \multicolumn{1}{r|}{1.074} & \multicolumn{1}{r}{\textbf{0.469}} & \multicolumn{1}{r}{0.597} & \multicolumn{1}{r}{0.554} & \multicolumn{1}{r}{0.539} & \multicolumn{1}{r|}{0.662} & \multicolumn{1}{r}{0.460} & \multicolumn{1}{r}{\underline{\textbf{0.320}}} & \multicolumn{1}{r}{0.576} & \multicolumn{1}{r}{0.387} & \multicolumn{1}{r}{0.597} \\
    \end{tabular}
    \vspace{-0.2cm}
\end{center}
\end{table*}

\textbf{Dataset.} We deploy the dataset proposed by Stahlke et al.~\cite{stahlke2022transfer} using a 5G-compatible SDR downlink system with six access points (APs). Each AP is equipped with a single antenna and operates with a bandwidth of 100\,MHz at a carrier frequency of 3.75\,GHz, transmitting at a power of 20\,dBm. The user equipment captures bursts at a frequency of 100\,Hz. To obtain accurate ground truth position data, a Nikon iGPS optical reference system, characterized by a mean absolute error (MAE) of less than $1\,mm$, was employed~\cite{rainer_mautz}. A large-scale measurement campaign was conducted, capturing scenarios with varying propagation conditions to represent different levels of channel complexity. These include a pure LoS environment devoid of objects and five mixed scenarios with varying object placements: two deterministic sites (DS1 and DS2) and three realistic sites (RS3, RS4, and RS5). For each site, a 1.5-hour recording was conducted, resulting in approximately 3 million CIRs along with corresponding ToA measurements. The deterministic sites are characterized by challenging propagation conditions, including absorber walls that induce systematic and deterministic blockages as well as dense multipath effects. In the realistic sites, additional industrial elements such as moving vehicles, forklifts, and elevated platforms were introduced alongside the absorber walls, simulating dynamic industrial environments and creating realistic variations in the propagation environment. These elements contribute to the generation of new multipaths and distinct radio fingerprints~\cite{stahlke2022transfer}.

\setlength{\intextsep}{6pt}
\setlength{\columnsep}{12pt}
\begin{wraptable}{R}{2.6cm}
    \begin{minipage}[b]{1.0\linewidth}
    \begin{center}
    \setlength{\tabcolsep}{2.0pt}
    \vspace{-0.2cm}
    \caption{Number of samples.}
    \label{table_dataset_samples}
    \footnotesize \begin{tabular}{ p{0.5cm} | p{0.5cm} }
    \multicolumn{1}{c|}{\textbf{Dataset}} & \multicolumn{1}{c}{\textbf{\# Samples}} \\ \hline
    \multicolumn{1}{c|}{RS3} & \multicolumn{1}{r}{280,817} \\
    \multicolumn{1}{c|}{RS4} & \multicolumn{1}{r}{170,398} \\
    \multicolumn{1}{c|}{RS5} & \multicolumn{1}{r}{115,374} \\
    \multicolumn{1}{c|}{DS1} & \multicolumn{1}{r}{269,509} \\
    \multicolumn{1}{c|}{DS2} & \multicolumn{1}{r}{284,241} \\
    \end{tabular}
    \end{center}
    \end{minipage}
\end{wraptable}
\textbf{DIL.} Initially, the model is trained on a primary dataset using randomly initialized weights. Subsequently, incremental fine-tuning is performed on various datasets reflecting environmental changes, utilizing DIL techniques to mitigate catastrophic forgetting of previously learned domains. Table~\ref{table_dataset_overview} provides an overview of the training, adaptation, and test datasets. The models are evaluated across all five datasets: RS3, RS4, RS5, DS1, and DS2. Table~\ref{table_dataset_samples} provides a summary of the number of samples, clearly indicating the necessity of the reduction.

\section{Evaluation}
\label{label_evaluation}

\textbf{Hardware \& Training Setup.} For all experiments, we use Nvidia Tesla V100-SXM2 GPUs with 32 GB VRAM equipped with Core Xeon CPUs and 192 GB RAM. The vanilla Adam optimizer is employed with a multi-step learning rate of $0.001$ and a batch size of $16$. The models are trained for 50 epochs on the initial dataset and for only five epochs on the adaptation datasets. Each model is trained five times, and the MAE mean and standard variance of the positioning error are reported.

\begin{figure}[!t]
    \centering
	\begin{minipage}[t]{0.486\linewidth}
        \centering
        \includegraphics[trim=7 7 7 7, clip, width=1.0\linewidth]{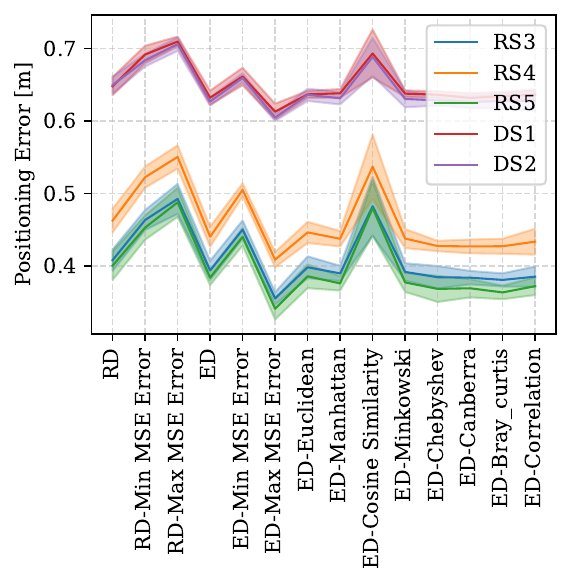}
        \subcaption{EWC.}
        \label{figure_distance_metrics1}
    \end{minipage}
    \hfill
	\begin{minipage}[t]{0.498\linewidth}
        \centering
        \includegraphics[trim=7 7 7 7, clip, width=1.0\linewidth]{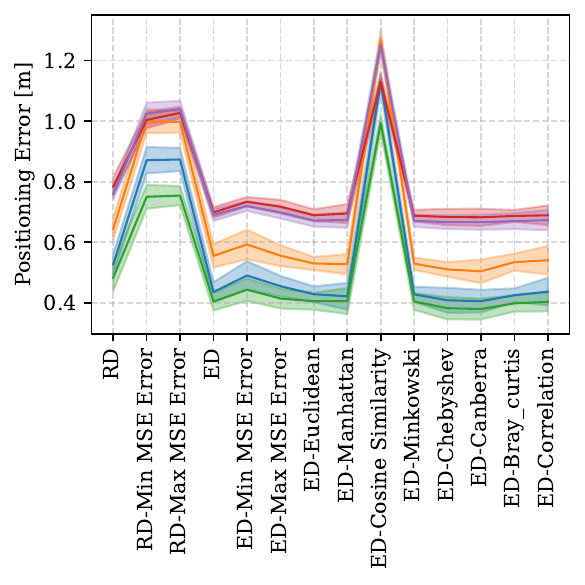}
        \subcaption{LwF.}
        \label{figure_distance_metrics2}
    \end{minipage}
    \caption{Evaluation of distance metrics for similarity-aware sample selection for the adaptation set RS3 $\rightarrow$ RS4 $\rightarrow$ RS5. RD: randomly distributed, ED: equally distributed. $N=50$.}
    \label{figure_distance_metrics}
\end{figure}

\begin{figure*}[!t]\captionsetup[subfigure]{font=scriptsize}
    \centering
	\begin{minipage}[t]{0.137\linewidth}
        \centering
        \includegraphics[trim=10 11 10 11, clip, width=1.0\linewidth]{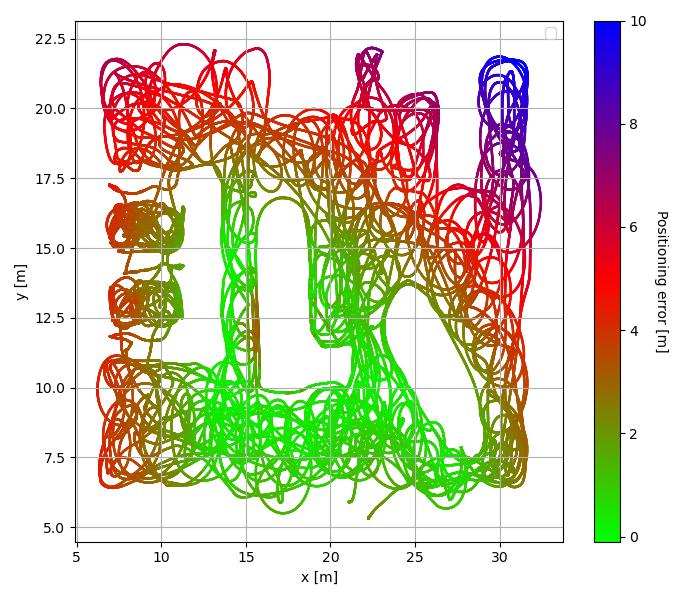}
        \subcaption{LwF, no exemplars, $1.241\,m$.}
        \label{figure_trajectories1}
    \end{minipage}
    \hfill
	\begin{minipage}[t]{0.137\linewidth}
        \centering
        \includegraphics[trim=10 11 10 11, clip, width=1.0\linewidth]{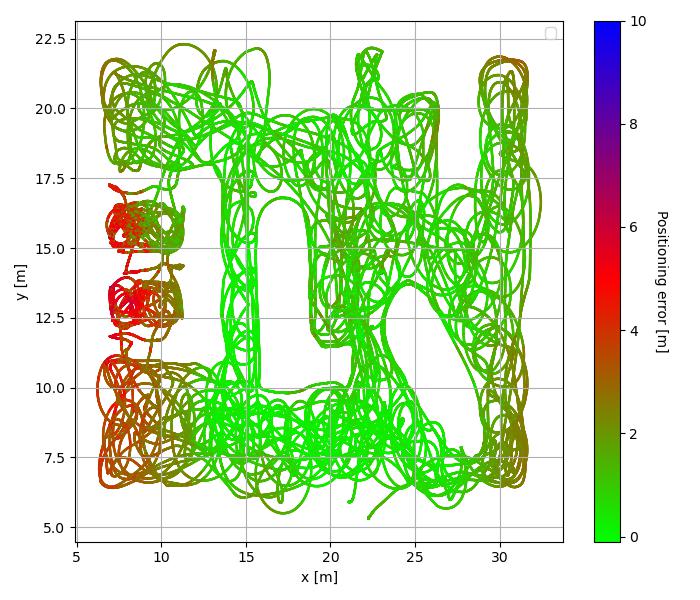}
        \subcaption{Finetune, 50 exemplars, $0.789\,m$.}
        \label{figure_trajectories2}
    \end{minipage}
    \hfill
	\begin{minipage}[t]{0.137\linewidth}
        \centering
        \includegraphics[trim=10 11 10 11, clip, width=1.0\linewidth]{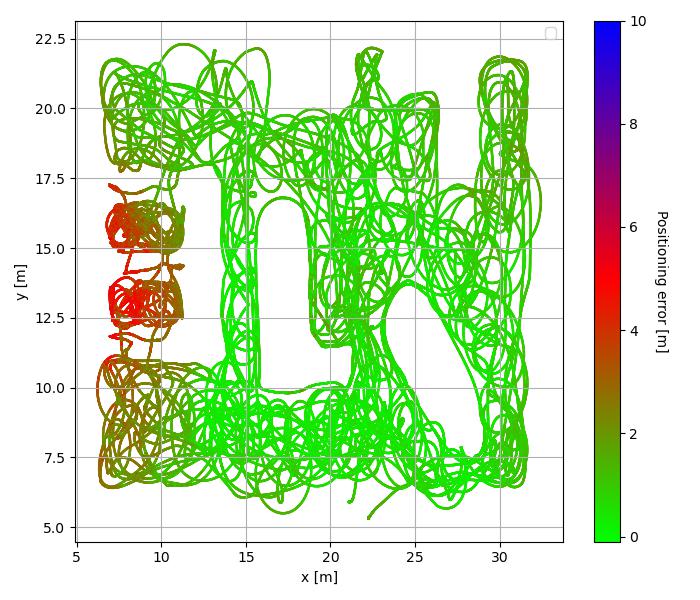}
        \subcaption{EWC, 50 exemplars, $0.454\,m$.}
        \label{figure_trajectories3}
    \end{minipage}
    \hfill
	\begin{minipage}[t]{0.137\linewidth}
        \centering
        \includegraphics[trim=10 11 10 11, clip, width=1.0\linewidth]{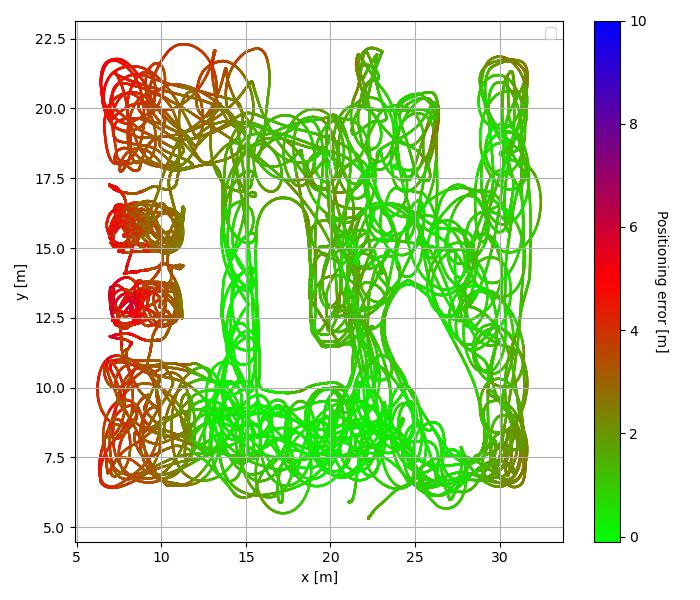}
        \subcaption{SI, 50 exemplars, $0.692\,m$.}
        \label{figure_trajectories5}
    \end{minipage}
	\begin{minipage}[t]{0.137\linewidth}
        \centering
        \includegraphics[trim=10 11 10 11, clip, width=1.0\linewidth]{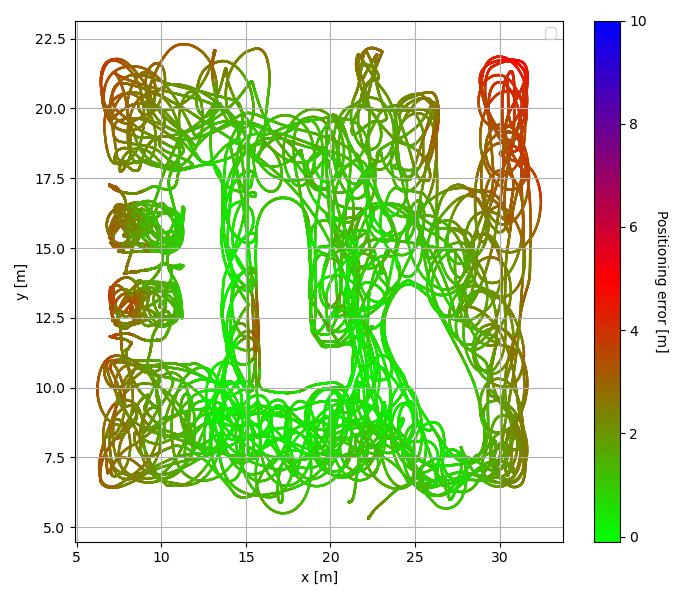}
        \subcaption{LwF, random sample selection, $0.714\,m$.}
        \label{figure_trajectories7}
    \end{minipage}
    \hfill
	\begin{minipage}[t]{0.137\linewidth}
        \centering
        \includegraphics[trim=10 11 10 11, clip, width=1.0\linewidth]{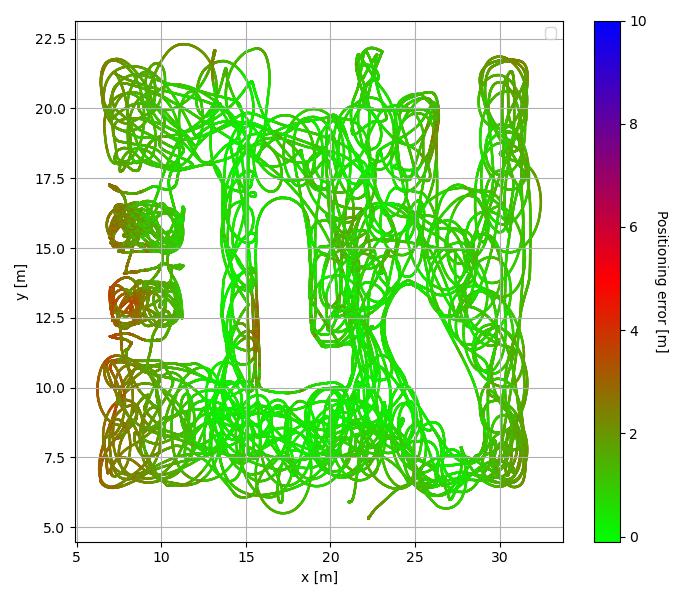}
        \subcaption{LwF, equally distributed , $0.526\,m$.}
        \label{figure_trajectories8}
    \end{minipage}
    \hfill
	\begin{minipage}[t]{0.137\linewidth}
        \centering
        \includegraphics[trim=10 11 10 11, clip, width=1.0\linewidth]{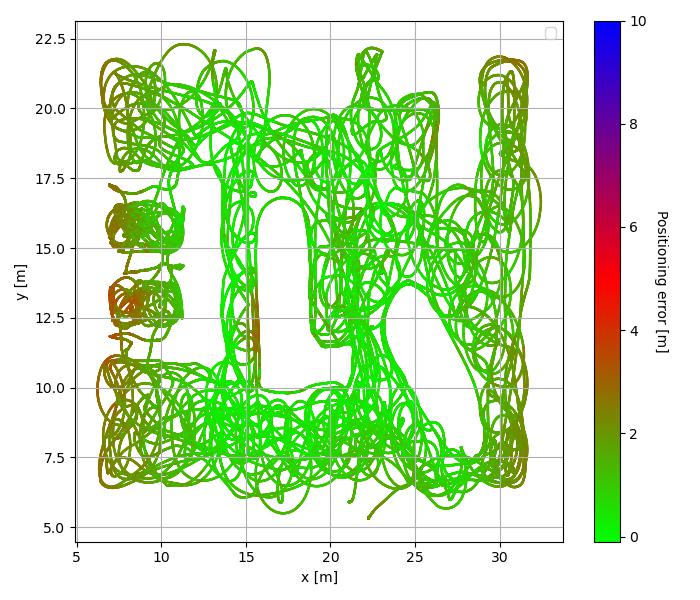}
        \subcaption{LwF, highest error-dependent, $0.568\,m$.}
        \label{figure_trajectories9}
    \end{minipage}
    \hfill
	\begin{minipage}[t]{0.137\linewidth}
        \centering
        \includegraphics[trim=10 11 10 11, clip, width=1.0\linewidth]{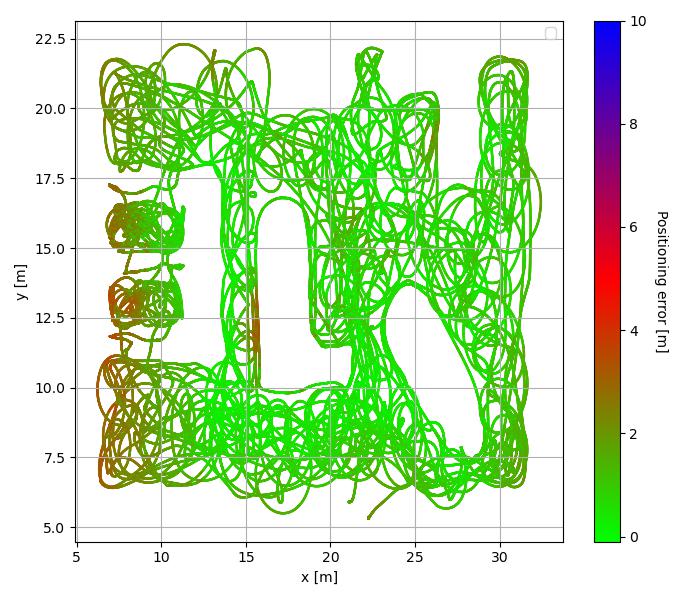}
        \subcaption{LwF, similarity (Manhattan), $0.500\,m$.}
        \label{figure_trajectories10}
    \end{minipage}
    \hfill
	\begin{minipage}[t]{0.137\linewidth}
        \centering
        \includegraphics[trim=10 11 10 11, clip, width=1.0\linewidth]{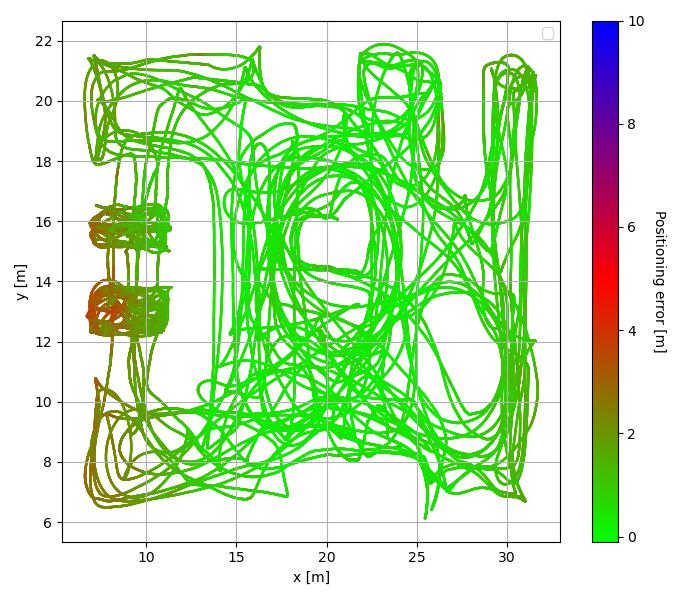}
        \subcaption{RS3, $0.366\,m$.}
        \label{figure_trajectories11}
    \end{minipage}
	\begin{minipage}[t]{0.137\linewidth}
        \centering
        \includegraphics[trim=10 11 10 11, clip, width=1.0\linewidth]{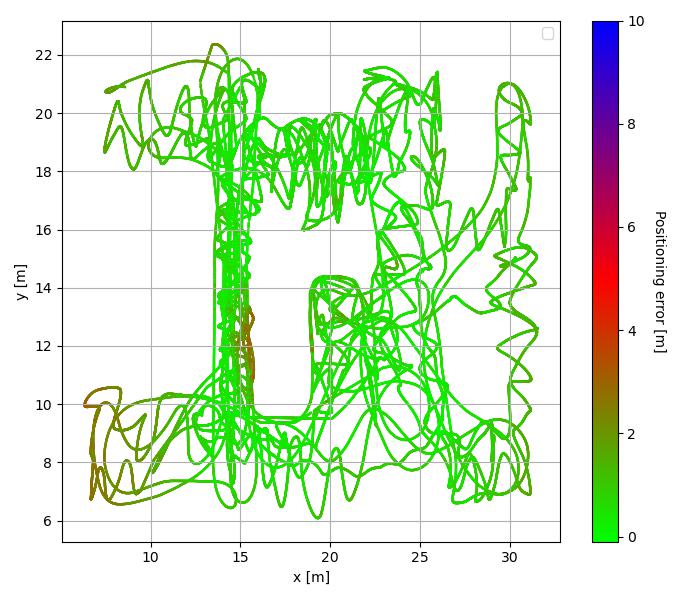}
        \subcaption{RS5, $0.387\,m$.}
        \label{figure_trajectories12}
    \end{minipage}
    \hfill
	\begin{minipage}[t]{0.137\linewidth}
        \centering
        \includegraphics[trim=10 11 10 11, clip, width=1.0\linewidth]{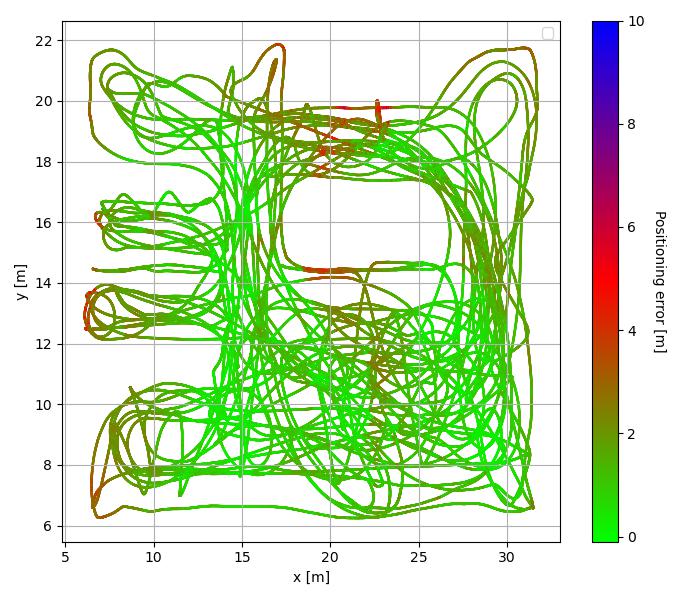}
        \subcaption{DS1, $0.690\,m$.}
        \label{figure_trajectories13}
    \end{minipage}
    \hfill
	\begin{minipage}[t]{0.137\linewidth}
        \centering
        \includegraphics[trim=10 11 10 11, clip, width=1.0\linewidth]{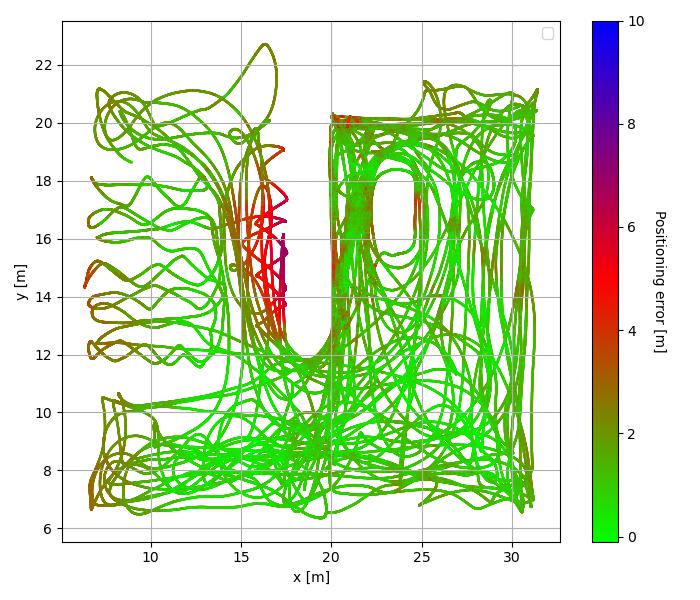}
        \subcaption{DS2, $0.678\,m$.}
        \label{figure_trajectories14}
    \end{minipage}
    \hfill
	\begin{minipage}[t]{0.137\linewidth}
        \centering
        \includegraphics[trim=10 11 10 11, clip, width=1.0\linewidth]{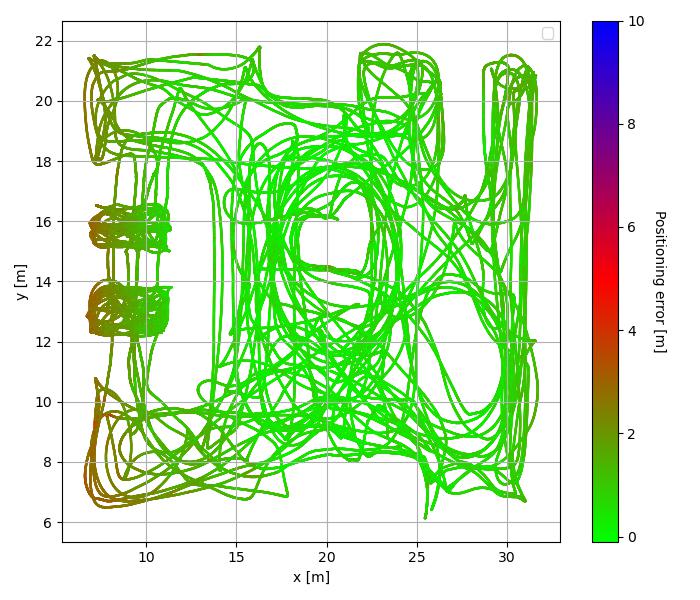}
        \subcaption{RS3 $\rightarrow$ DS1, test RS3, $0.454\,m$.}
        \label{figure_trajectories15}
    \end{minipage}
    \hfill
	\begin{minipage}[t]{0.137\linewidth}
        \centering
        \includegraphics[trim=10 11 10 11, clip, width=1.0\linewidth]{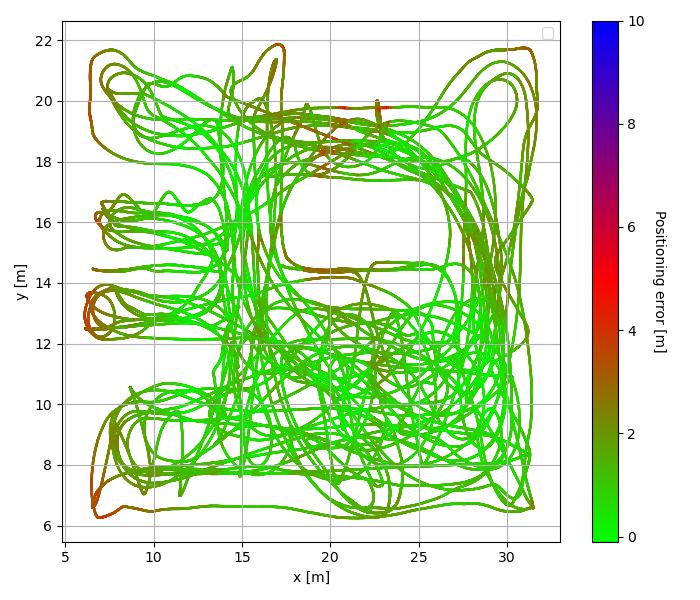}
        \subcaption{RS3 $\rightarrow$ DS1, test DS1, $0.602\,m$.}
        \label{figure_trajectories16}
    \end{minipage}
    \vspace{-0.15cm}
    \caption{Predicted trajectories and the corresponding errors in $m$ for no exemplars (a), all DIL methods with $N=50$ exemplars (b to e), and different sample selection methods (e to h) for the adapation from domain RS3 $\rightarrow$ RS4 $\rightarrow$ RS5 and evaluated on RS4. Evaluation of the best performing technique LwF with similarity-based sample selection (Manhattan distance) with $N=50$ exemplars on all test domains (i to l) and for the adaptation RS3 $\rightarrow$ DS1 (m and n).}
    \label{figure_trajectories}
    \vspace{-0.25cm}
\end{figure*}

\textbf{DIL Evaluation.} Table~\ref{table_dataset_results} summarizes all evaluation results, while Figure~\ref{figure_trajectories} presents the corresponding evaluation trajectories. For $N=0$, EWC outperforms all other DIL methods by effectively retaining knowledge of older tasks, thus mitigating catastrophic forgetting, highlighting the weakness of Finetune (FT). For $N \in \{50, 100\}$, LwF achieves the lowest positioning error, benefiting from the additional examples that enhance performance, as shown in Figures~\ref{figure_trajectories2} through \ref{figure_trajectories5}. However, for $N=200$, all methods, including FT, demonstrate significant improvement, with EWC and LwF yielding the lowest errors. When sufficient exemplars are available, these methods effectively prevent catastrophic forgetting. PNN performs well in scenarios with shared characteristics but struggles significantly when faced with diverse scenarios containing different objects, highlighting its limitation in generalizing to new, unseen scenarios. While performance remains highly robust for small changes ($< 0.4\,m$, as seen in Figures~\ref{figure_trajectories9} and \ref{figure_trajectories10}), error increases with larger changes ($> 0.6\,m$, as shown in Figures~\ref{figure_trajectories11} and \ref{figure_trajectories12}), although the methods still exhibit robustness to these changes.

\textbf{Evaluation of Sample Selection.} Figure~\ref{figure_distance_metrics} presents the results for sample selection methods. ED selection significantly outperforms RD selection for both EWC and LwF, prompting further exploration of distance metrics for ED. For EWC, the ED $\text{MSE}_{\text{min}}$-based selection ($0.365\,m$) and distance-based methods ($0.391\,m$) achieve the lowest error. Similarly, for LwF, the Chebyshev distance yields the best performance ($0.360\,m$). As cosine similarity proves ineffective, the choice of distance metric is crucial. Both Canberra and Manhattan distances perform comparably well to Chebyshev. For example, the performance improvement is evident in the right trajectory of Figure~\ref{figure_trajectories10}.

\begin{figure}[!t]
    \centering
    \vspace{-0.2cm}
	\begin{minipage}[t]{0.325\linewidth}
        \centering
        \includegraphics[trim=10 10 10 10, clip, width=1.0\linewidth]{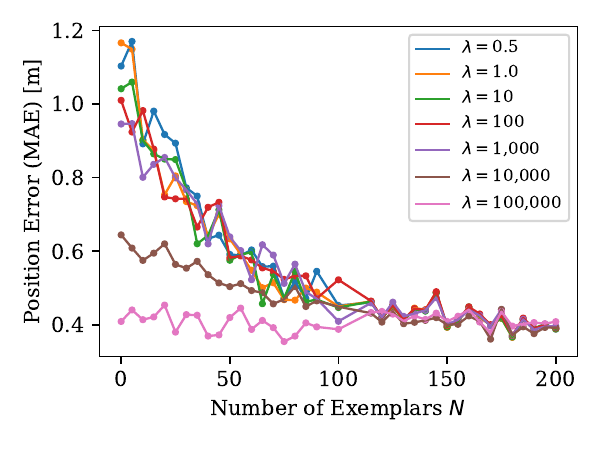}
        \subcaption{EWC.}
        \label{figure_hyperparameters1}
    \end{minipage}
    \hfill
	\begin{minipage}[t]{0.325\linewidth}
        \centering
        \includegraphics[trim=10 10 10 10, clip, width=1.0\linewidth]{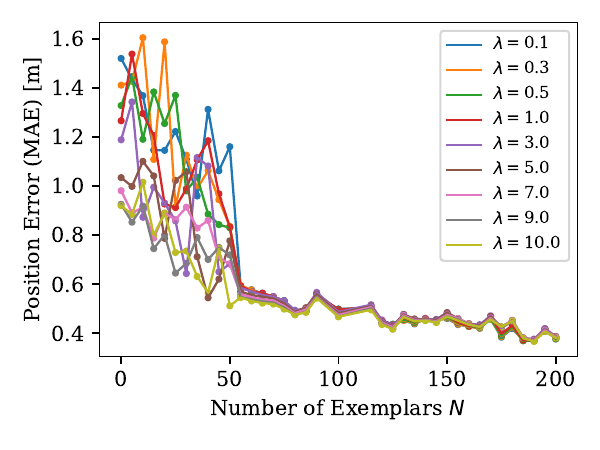}
        \subcaption{LwF.}
        \label{figure_hyperparameters2}
    \end{minipage}
    \hfill
	\begin{minipage}[t]{0.325\linewidth}
        \centering
        \includegraphics[trim=10 10 10 10, clip, width=1.0\linewidth]{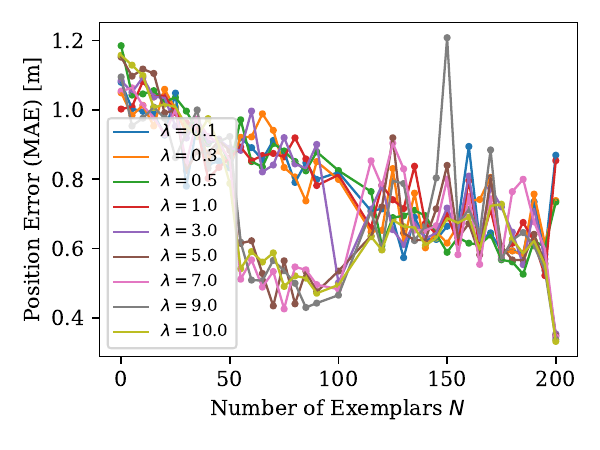}
        \subcaption{SI.}
        \label{figure_hyperparameters3}
    \end{minipage}
    \caption{Hyperparameter search of $\lambda$ for the adaptation RS3 $\rightarrow$ RS4 on the RS3 test dataset for EWC, LwF, and SI.}
    \label{figure_hyperparameters}
\end{figure}

\textbf{Hyperparameter Searches \& Size of Exemplar Set.} Figure~\ref{figure_hyperparameters} presents the evaluation of hyperparameter searches for $\lambda$ and the selection of the exemplar set size in the methods EWC, LwF, and SI. In EWC and LwF, $\lambda$ regulates the strength of the regularization that mitigates catastrophic forgetting by controlling the balance between retaining knowledge from previous tasks and acquiring new ones, penalizing significant deviations of important weights. A higher value of $\lambda = 100,000$ results in the most consistent positioning error of approximately $0.4\,m$. For small exemplar sets ($N<100$), the positioning error increases with smaller $\lambda$ values, whereas for larger exemplar sets ($N>100$), $\lambda$ becomes less relevant. For LwF, the search focused on smaller $\lambda$ values ranging between 0.1 and 10.0. Similar to EWC, the weighting of $\lambda$ becomes less significant for larger exemplar sets ($N>50$), while for smaller exemplar sets, higher values ($\lambda = 10.0$) yield optimal performance. In SI, $\lambda$ controls the regularization strength that constrains modifications to critical synapses. For exemplar set sizes greater than 50, optimal results are obtained with $\lambda$ values of 5.0, 7.0, 9.0, and 10.0, achieving a positioning error of $0.4\,m$. Careful tuning of $\lambda$ and the selection of an appropriate exemplar set size are essential to balancing the minimization of catastrophic forgetting with the effective acquisition of new domains.

\begin{figure}[!t]
    \centering
    \vspace{-0.2cm}
	\begin{minipage}[t]{0.49\linewidth}
        \centering
        \includegraphics[trim=7 7 7 7, clip, width=1.0\linewidth]{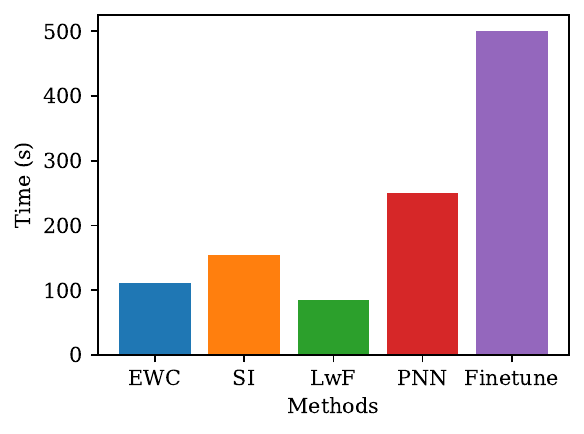}
    \end{minipage}
    \hfill
	\begin{minipage}[t]{0.49\linewidth}
        \centering
        \includegraphics[trim=7 7 7 7, clip, width=1.0\linewidth]{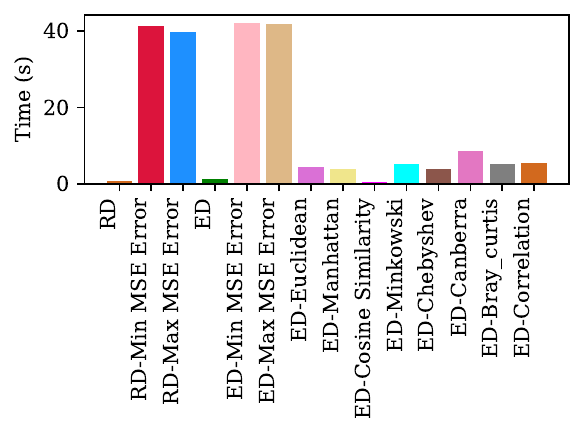}
    \end{minipage}
    \caption{Evaluation of training times [in $s$] for DIL methods.}
    \label{figure_training_times}
\end{figure}

\textbf{Computation Times.} Figure~\ref{figure_training_times} presents the training times for DIL and sample selection methods over five epochs on the adaptation set. FT, which trains the entire network, requires $500\,s$, while PNN expands the network with each new task ($250\,s$). EWC necessitates $100\,s$ for training, with an additional $10\,s$ for computing the Fisher matrix, whereas SI requires $155\,s$ to track parameter importance. LwF, leveraging knowledge distillation, trains more efficiently in $85\,s$. These results highlight the efficiency of LwF and EWC in mitigating forgetting in DIL tasks. Additionally, we analyze the sample selection methods. RD and ED methods are computationally efficient, as they do not require additional computations, though they yield lower performance. MSE-based methods, which prioritize samples where the model is most confident or struggles the most, demand the highest computation time (approx.~$40\,s$). In contrast, our distance-based methods are significantly faster due to the simplicity of distance calculations.
\section{Conclusion}
\label{label_conclusion}

For 5G indoor-based localization, we proposed a DIL method designed to continuously adapt to dynamic environmental changes. Our EWC and LwF-based sample selection methods resulted in a significant reduction in exemplar sizes for rapid training, while achieving errors of $0.261\,m$ for small changes and $0.266\,m$ for larger changes. Our framework collects data only from dynamic regions, retaining static knowledge to reduce overhead and improve scalability.

\section*{Acknowledgments}
\small This work has been carried out within the DARCII project, funding code 50NA2401, sponsored by the German Federal Ministry for Economic Affairs and Climate Action (BMWK) and supported by the German Aerospace Center (DLR), the Bundesnetzagentur (BNetzA), and the Federal Agency for Cartography and Geodesy (BKG).

\bibliography{ICL2025}
\bibliographystyle{IEEEtran}

\end{document}